\documentclass[sigconf]{acmart}
\AtBeginDocument{%
  }

\setcopyright{acmlicensed}
\copyrightyear{2026}
\acmYear{2026}
\acmDOI{XXXXXXX.XXXXXXX}
\acmConference[CAIS '26]{2026 Call for System Demonstrations}{TBD}{TBD}
\acmISBN{978-1-4503-XXXX-X/2026/XX}

\begin{document}

\title{ClinicBot: A Guideline-Grounded Clinical Chatbot with Prioritized Evidence RAG and Verifiable Citations}

\author{Navapat Nananukul}
\email{nananuku@usc.edu}
\affiliation{%
  \institution{USC Information Sciences Institute}
  \city{Los Angeles}
  \state{California}
  \country{USA}
}

\author{Mayank Kejriwal}
\email{kejriwal@isi.edu}
\affiliation{%
  \institution{USC Information Sciences Institute}
  \city{Los Angeles}
  \state{California}
  \country{USA}
}

\renewcommand{\shortauthors}{Nananukul and Kejriwal}

\begin{abstract}
Clinical diagnosis requires answers that are accurate, verifiable, and explicitly grounded in official guidelines. While large language models excel at natural language processing, their tendency to hallucinate undermines their utility in high-stakes medical contexts where precision is essential. Existing retrieval-augmented generation (RAG) systems treat all evidence equally, producing noisy context and generic answers misaligned with clinical practice. We present ClinicBot, an AI system that translates guideline recommendations into trustworthy clinical support through three key advances: (1) structured extraction of clinical guidelines into semantic units (recommendations, tables, definitions, narrative) with explicit provenance, (2) evidence prioritization that ranks content by clinical significance and guideline structure rather than textual similarity, and (3) a web-based interface that presents concise, actionable answers with verifiable evidence. We will demonstrate ClinicBot using diabetes questions from real patients and an additional diabetes risk assessment tool that is faithful to the American Diabetes Association (ADA) \textit{Standards of Care in Diabetes (2025)}. The demonstration will illustrate how semantic knowledge extraction and hierarchical evidence ranking can reliably operate in a multi-agent setting to process complex clinical guidelines at scale.
\end{abstract}


\received{13 March 2026}
\received[revised]{13 March 2026}
\received[accepted]{13 March 2026}

\maketitle

\section{Introduction}

In recent years, the deployment of large language models (LLMs) for tasks demanding grounded reasoning has inspired significant ideas in retrieval-augmented generation (RAG), prompt engineering, and multi-agent systems \cite{lewis2020rag,wei2022cot,trivedi2023ircot}. However, LLMs have been limited to rigorous evaluation in \emph{high-assurance domains} such as clinical medicine, where reasoning must be accurate, verifiable, and explicitly grounded in authoritative guidelines \cite{thirunavukarasu2023large,singhal2025toward}. Trustworthy clinical AI requires not only high accuracy but also explainability and the ability to justify decisions with evidence \cite{future2024}. While RAG can ground answers in external sources, traditional RAG over lengthy clinical documents often treats evidence indiscriminately, producing noisy context and generic answers misaligned with clinical evidence hierarchies. Moreover, standard passage-level retrieval obscures the relationship between recommendations and supporting evidence, making it difficult for clinicians to verify whether answers faithfully reflect information in the guideline, even when the answers appear correct \cite{shi2023distracted,coussement2024explainable}.

In this work, we introduce \textsc{ClinicBot}, an AI system that enhances traditional RAG using semantic extraction and prioritized retrieval. Rather than treating the guideline as undifferentiated text for generic RAG, \textsc{ClinicBot} extracts and classifies guideline content into a \textit{structured knowledge base} containing recommendations, criteria tables, and narrative text. Each element is assigned with explicit clinical priority. In parallel, the system employs a \textit{two-step retrieval pipeline} that first routes a clinician query to a relevant guideline section or subsection, then retrieves evidence in strict priority order (recommendations $>$ insight tables and statistics $>$ supporting narrative). We demonstrate the system across two complementary use cases (guideline-grounded question answering and diabetes risk assessment) on the ADA Standards of Care in Diabetes 2025 \cite{ada2025}. Our design combines semantic knowledge extraction, question routing, evidence prioritization, and validation, to ground answers in authoritative guidelines and enforce full traceability from answer to source material. It is supported by a web-based interface developed following feedback from real doctors.

\section{Methodology}

\begin{figure*}[t]
\centering
\includegraphics[width=0.75\textwidth]{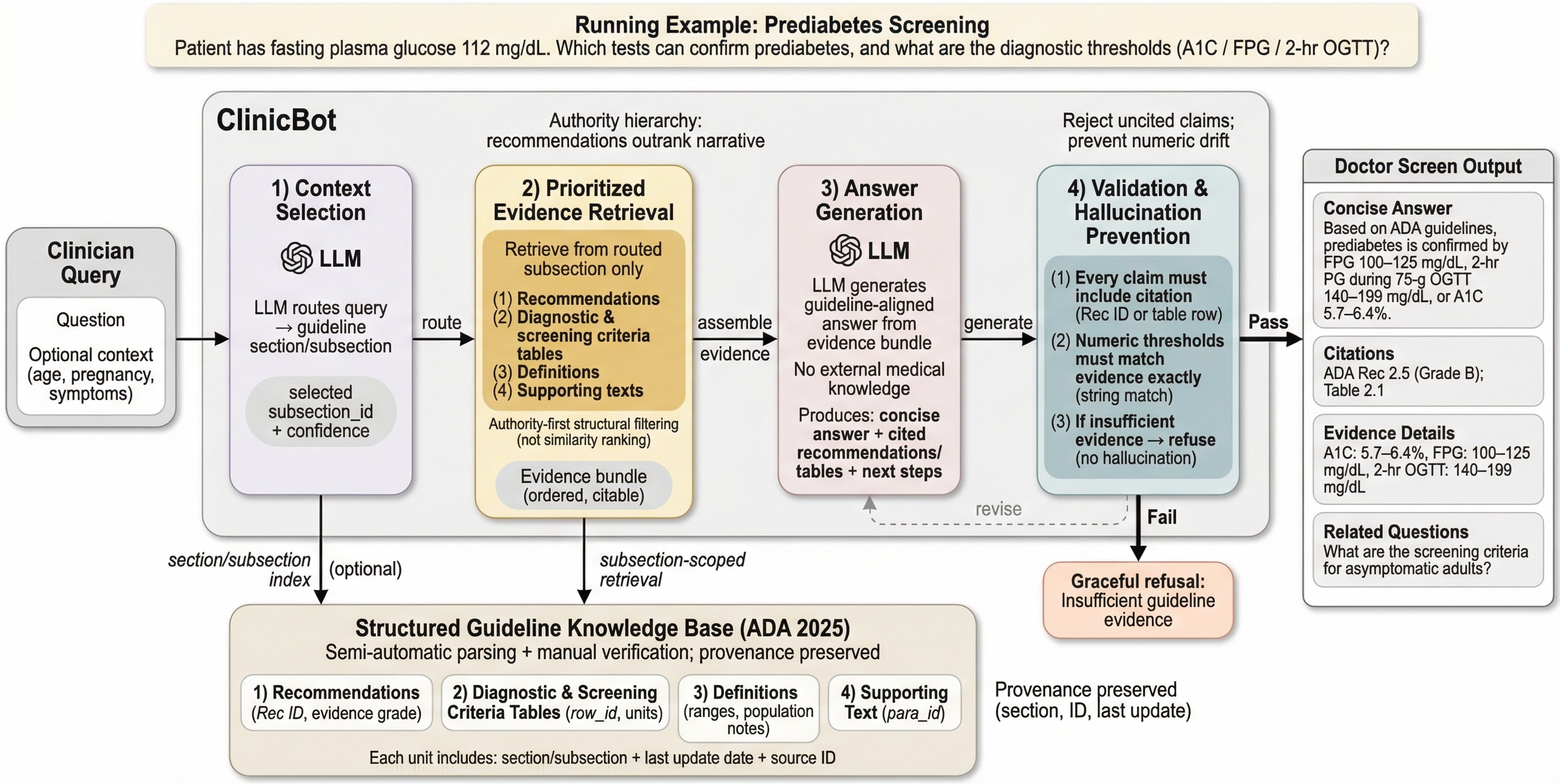}
\caption{\textit{Guideline-Structured RAG System Architecture.} The system routes clinician queries to relevant guideline subsections, retrieves evidence in priority order (recommendations first, then structured criteria tables, then supporting text), generates guideline-aligned concise answers with explicit citations, validates that all claims are grounded in retrieved evidence, and presents results via a two-part interface: \textit{Concise Answer} and \textit{Supporting Evidence}. The structured guideline knowledge base preserves source attribution and enables full traceability.}
\label{fig:architecture}
\end{figure*}

\noindent As shown in Figure \ref{fig:architecture}, ClinicBot improves upon generic RAG for clinical applications by decoupling retrieval into two complementary stages. Given a guideline document and a clinical question, we first construct a structured knowledge base that classifies guideline content by clinical priority—recommendations, tables, and narrative texts. Following evidence hierarchy principles, we rank content by clinical authority: grade-level recommendations and structured criteria tables take precedence over supporting narrative \cite{guyatt2011grade}. In the second stage, ClinicBot performs routing-then-retrieval, which maps the question to relevant guideline sections before assembling evidence. 

\paragraph{\textbf{Guidelines to knowledge base.}} Unlike generic RAG systems that operate over PubMed abstracts or web search results, we ground the current version of ClinicBot in a single authoritative source: the \textit{ADA Standards of Care in Diabetes---2025}, a peer-reviewed clinical guideline. This single-source grounding prioritizes answer quality and guideline fidelity over general knowledge from unfiltered LLMs or RAG systems. We explicitly classify content into three types, each with distinct retrieval priority: (1)~Recommendations with IDs and evidence grades, extracted as structured records; (2)~Diagnostic and screening criteria tables with numeric thresholds and classification ranges; (3)~Narrative context including rationale, population-specific guidance, and definitions. All content is extracted into JSON format using semi-automatic parsing and manual verification, with complete source attribution (guideline section, recommendation ID, date).

\paragraph{\textbf{Context selection.}} The pipeline begins with a routing step that maps the clinician's question to a relevant guideline section and subsection using an LLM-based few-shot classifier. The router classifies the query against the predefined section and subsection structure of the ADA guidelines (e.g., ``Diagnosis \& Screening'', ``Management of Diabetes in Pregnancy'', ``Screening and Prevention'') and selects the most appropriate section. This initial routing narrows the retrieval scope to a single, conceptually coherent section of the guideline, which improves specificity compared to searching the entire knowledge base.

\paragraph{\textbf{Prioritized evidence retrieval.}} Once the relevant section is identified, the retriever extracts all content units from that section and enforces a strict priority ordering that reflects clinical authority rather than similarity-based ranking. The retrieval order is: (1) recommendations, (2) diagnostic and screening criteria tables, and (3) supporting texts. This ordering is enforced structurally by parsing the JSON-formatted knowledge base and filtering content by type, rather than via post-hoc reranking. It ensures that the most clinically actionable and citable content always dominates the retrieved context. Recommendations and tables are always ranked above general narrative, reflecting their clinical authority and evidence grades. The system assembles an \textit{evidence bundle} containing all selected content, ordered by priority tier, which is then passed to the generation phase.

\paragraph{\textbf{Answer generation.}} At this step, we use an LLM to generate a response based on the evidence bundle assembled in the retrieval phase. The generator uses a prompt template that enforces a strict constraint: all claims must be grounded in retrieved content. Numeric thresholds, recommendations, and clinical next steps are derived directly from the knowledge base; no external medical knowledge is injected. The generated response includes: (1) A concise clinical answer (2--3 sentences) that is direct, actionable, and free of embedded citations (citations and evidence details are presented in a separate supporting section). (2) Source citations linking claims to source recommendations and diagnostic tables (e.g., ``Rec 2.2a (B)''). (3) Clinical recommendations — specific next steps extracted from authoritative guideline statements. (4) Evidence details — precise numeric thresholds, diagnostic ranges, and supporting criteria. (5) Related questions — follow-up prompts based on retrieved content to guide further exploration.

\paragraph{\textbf{Validation and hallucination prevention.}} Before an answer is presented to the user, an LLM-based validation module enforces three mandatory constraints to mitigate hallucinations—a known critical concern in clinical AI \cite{aljohani2025trustworthiness}: (1) Every claim must be supported by retrieved evidence and include a citation to a specific recommendation ID or table row number; claims without citations are rejected. (2) Any numeric thresholds cited (e.g., ``FPG 100--125 mg/dL'') must match retrieved values exactly via string matching, preventing subtle numeric drift. (3) If the routed section does not contain sufficient evidence for the question, the system declines to generate unsupported text and instead returns a graceful refusal message (``Insufficient guideline evidence for this question'') rather than hallucinating. This design prevents the system from generating plausible-sounding but ungrounded claims and ensures all answers are fully traceable to source material.


\section{ClinicBot: Demonstration, Implementation, and Evaluation}

\noindent This section presents the core functionality and user experience of ClinicBot through two complementary use case demonstrations: guideline-grounded medical question answering and diabetes risk assessment. We then report quantitative evaluation results showing how effectively the system translates complex clinical guidelines into accurate, evidence-backed responses for practicing clinicians.

\noindent During the demonstration and review process, users can interact directly with ClinicBot through an intuitive web interface.\footnote{ClinicBot is  available at: \url{https://shorturl.at/OoZvT}} users can submit medical questions/scenarios and observe the system's response pipeline in real time. This end-to-end interaction demonstrates the core value of ClinicBot's architecture—attendees can verify that answers are directly grounded in guideline sources.

\subsection{Use Case 1: Medical Question Answering}

\begin{figure*}[t]
\centering
\includegraphics[width=0.83\textwidth]{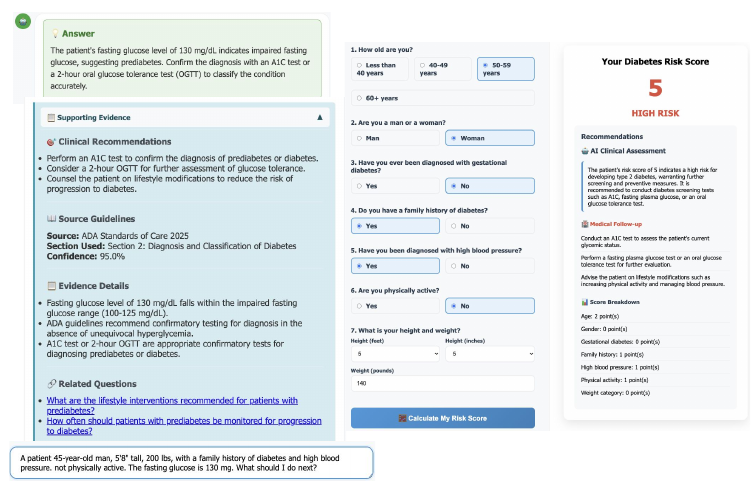}
\caption{ClinicBot's interfaces demonstrate guideline-grounded clinical support. \textit{Left}: Medical question answering interface, where clinicians submit patient queries and receive two-part responses: a concise answer (highlighted in green) followed by expandable supporting evidence with citations and clinical rationale. \textit{Right}: Diabetes risk assessment interface, where clinicians input patient information and ClinicBot computes risk scores with clinical interpretation and actionable recommendations.}
\label{fig:usecase}
\end{figure*}

\noindent Figure \ref{fig:usecase} (left) illustrates the primary use case of ClinicBot in which a clinician inputs a patient scenario to obtain guideline-grounded answers. A clinician enters a detailed patient case via the chat interface: ``A 45-year-old man, 5'8'' tall, 200 lbs, with a family history of diabetes and high blood pressure, is not physically active. His fasting glucose is 130 mg/dL. What should I do next?'' The system analyzes this query, identifies relevant clinical concepts (impaired fasting glucose, prediabetes risk, comorbidities), and routes the question to the appropriate guideline sections. This allows ClinicBot to select Section 2: Diagnosis and Classification of Diabetes as the most relevant context. In this case, the system retrieves from section 2 in this order: (1)~Recommendation 2.1 stating that FPG 100--125 mg/dL defines impaired fasting glucose (IFG), indicating prediabetes (Evidence Grade: A); (2)~Diagnostic criteria table specifying A1C 5.7--6.4\% or 2-hour postprandial glucose 140--199 mg/dL during OGTT also indicate prediabetes; (3)~Narrative text explaining risk factors including family history and physical inactivity.

\noindent The system then generates a two-part response: (1)~\textbf{Concise answer} synthesizes the retrieved evidence into a direct clinical statement: ``FPG of 130 mg/dL indicates impaired fasting glucose (prediabetes). Confirm diagnosis with A1C test or 2-hour OGTT to assess progression risk.'' This answer cites specific recommendations (e.g., ``Rec 2.1a'') and numeric thresholds matched exactly from source tables. (2)~\textbf{Supporting evidence} contains: clinical next steps extracted from guideline statements, source citations linking each claim to recommendation IDs or table rows, evidence details with numeric thresholds and diagnostic ranges, and related follow-up questions.

\subsection{Use Case 2: Diabetes Risk Assessment}

\noindent ClinicBot includes a complementary risk assessment tool that follows the ADA Diabetes Risk Test. The tool employs a questionnaire interface where clinicians or patients can input seven key items: (1)~age group, (2)~biological sex, (3)~history of gestational diabetes or gestational hypertension, (4)~family history of diabetes among parents or siblings, (5)~personal history of high blood pressure (currently on treatment or untreated), (6)~physical activity level (frequency of exercise per week), and (7)~anthropometric measurements (height and weight for BMI calculation).

\noindent Upon submission, the system computes a cumulative risk score based on the official ADA framework, where each risk factor contributes a specific number of points. The system displays the risk assessment results across four complementary components, each grounded in ADA guidelines: (1)~\textbf{Risk score and category}: A prominent display showing the computed total score and corresponding risk level (e.g., ``Score: 7 (Increased Risk)''), immediately signaling the patient's classification. (2)~\textbf{Clinical interpretation}: Evidence-based explanation of what the score means for the patient's diabetes risk trajectory and the clinical implications of their risk category. (3)~\textbf{Actionable recommendations}: Guideline-based next steps specific to the patient's risk level—for example, high-risk patients are advised to ``schedule fasting glucose or HbA1c testing within the next 1--3 months,'' while increased-risk patients are counseled on lifestyle modifications (increased physical activity, weight reduction) and the timeline for screening. (4)~\textbf{Score breakdown}: A detailed itemization of point contributions from each risk factor.

\subsection{System Implementation}

\paragraph{\textbf{Replication details.}} The demonstration system is implemented as a web application with a Python Flask backend and a single-page frontend. The backend components are implemented in Python~3.10+ using Flask for HTTP routing and request handling, \texttt{flask-cors} for cross-origin access, and \texttt{requests} for LLM API calls. All LLM calls are made via the OpenRouter API using GPT-4o \cite{openai2024gpt4}. We use low temperatures (typically $T{=}0.1$) for section routing and answer generation. We use LlamaIndex \cite{llamaindex2023} for efficient document indexing and retrieval optimization. To enable reproducible local deployment, users can clone the source code, install the required libraries, configure the openrouter API key variable, and run a single startup script.

\paragraph{\textbf{Knowledge base construction details.}} We extract content into a self-contained JSON file: recommendations are identified via regex patterns matching IDs and evidence grades, tables are extracted and normalized, and narrative blocks are grouped by topic headings. All components are extracted using \texttt{PyPDF2} and \texttt{pdfplumber}. Each unit is annotated with section, ID, and page number, then verified manually against the original PDF. The JSON is loaded into memory at initialization to construct a retrieval catalog indexed by section and content type. For embedding-based retrieval variants, we build a LlamaIndex \texttt{VectorStoreIndex} using OpenAI's \texttt{text-embedding-3-small} model for similarity pre-filtering before prioritized retrieval.

\noindent 

\subsection{Evaluation Results}

\noindent To assess ClinicBot's clinical accuracy and completeness, we evaluated the system against 30 manually curated diabetes-related questions sourced from two independent data sources: (1) the Hugging Face Diabetes QA dataset, providing expert-authored ground truth answers, and (2) the diabetes.co.uk medical webboard, where we extracted clinician- and expert-annotated responses as validation references. Our evaluation covers diagnostic criteria, medication management, lifestyle interventions, complications monitoring, and patient education. ClinicBot achieves 96\% combined accuracy across the evaluation set, demonstrating strong capability in translating guideline recommendations into clinically accurate responses.

\noindent This 96\% accuracy breaks down as follows: 19 of 30 responses (63\%) are fully correct, 10 of 30 responses (33\%) demonstrate correct understanding with minor incomplete details, and 1 of 30 responses (3\%) is incorrect. The minor incomplete cases typically involved the omission of non-critical details (such as secondary medication options) or less commonly referenced monitoring parameters, while maintaining core clinical guidance that remains actionable for practicing clinicians. These results demonstrate that ClinicBot's guideline-grounded architecture, with structured knowledge extraction and evidence prioritization, effectively operationalizes complex ADA recommendations into accurate, clinically substantive responses.

\section{Related Work}

\noindent \textbf{Retrieval-Augmented Generation and Clinical Decision Support.} Retrieval-augmented generation (RAG) grounds generative AI in external knowledge, addressing the hallucination risk endemic to LLMs in high-stakes domains like healthcare \cite{lewis2020rag, huang2025survey}. Classical RAG systems retrieve and rank passages by similarity to the query \cite{karpukhin2020dpr}, a strategy that works well for open-domain tasks but fails in specialized domains where clinical authority matters more than textual overlap. Recent medical RAG systems like MedRAG \cite{zhao2025medrag} enhance retrieval by incorporating knowledge graphs, while others employ domain-specific embeddings like MedCPT \cite{jin2023medcpt} to improve relevance. However, these approaches typically treat all retrieved evidence equally, whereas effective clinical decision support requires integrating evidence-based guidelines into actionable systems organized by evidence grade \cite{guyatt2011grade, kwan2020meta}. Well-designed CDSS that surface evidence justifications improve clinician adherence and reasoning \cite{chen2023harnessing}, yet the challenge lies in translating guideline knowledge into trustworthy computational delivery with full traceability.

\noindent \textbf{Medical AI Safety and Explainability.} Large language models have demonstrated impressive performance on medical knowledge tasks \cite{singhal2023med, singhal2025toward}, achieving USMLE-level reasoning through instruction tuning and prompt engineering. However, translating this capability into trustworthy clinical decision support requires more than raw reasoning ability—it demands explicit evidence grounding, validation against clinical guidelines, and explainability to clinicians. Recent work emphasizes the importance of hallucination prevention \cite{aljohani2025trustworthiness} and human-in-the-loop verification in high-assurance settings. Clinician-facing interfaces that surface source material and justifications alongside recommendations are increasingly emphasized in the literature \cite{thirunavukarasu2023large}. ClinicBot contributes to this direction by combining guideline-grounded retrieval with automated validation constraints that enforce answer traceability.

\bibliographystyle{ACM-Reference-Format}
\bibliography{references}

@inproceedings{lewis2020rag,
  title = {Retrieval-augmented generation for knowledge-intensive NLP tasks},
  author = {Lewis, Patrick and Perez, Ethan and Piktus, Aleksandra and Petroni, Fabio and Karpukhin, Vladimir and Goyal, Naman and K{\"u}ttler, Heinrich and Lewis, Mike and Yih, Wen-tau and Rockt{\"a}schel, Tim and Riedel, Sebastian and Kiela, Douwe},
  booktitle = {Advances in Neural Information Processing Systems},
  volume = {33},
  pages = {9459--9474},
  year = {2020}
}

@inproceedings{wei2022cot,
  title = {Chain-of-thought prompting elicits reasoning in large language models},
  author = {Wei, Jason and Wang, Xuezhi and Schuurmans, Dale and Bosma, Maarten and Ichter, Brian and Xia, Fei and Chi, Ed and Le, Quoc and Zhou, Denny},
  booktitle = {Advances in Neural Information Processing Systems},
  volume = {35},
  pages = {24824--24837},
  year = {2022}
}

@inproceedings{trivedi2023ircot,
  title = {Interleaving retrieval with chain-of-thought reasoning for knowledge-intensive multi-step questions},
  author = {Trivedi, Harsh and Balasubramanian, Niranjan and Khot, Tushar and Sabharwal, Ashish},
  booktitle = {Proceedings of the 61st Annual Meeting of the Association for Computational Linguistics},
  pages = {10014--10037},
  year = {2023}
}

@article{thirunavukarasu2023large,
  title = {Large language models in medicine},
  author = {Thirunavukarasu, Arun James and Ting, Darren Shu Jeng and Elangovan, Kabilan and Gutierrez, Laura and Tan, Ting Fang and Ting, Daniel Shu Wei},
  journal = {Nature Medicine},
  volume = {29},
  number = {8},
  pages = {1930--1940},
  year = {2023}
}

@article{singhal2025toward,
  title = {Toward expert-level medical question answering with large language models},
  author = {Singhal, Karan and Tu, Tao and Gottweis, Juraj and Sayres, Rory and Wulczyn, Ellery and Amin, Mohamed and Hou, Le and Clark, Kevin and Pfohl, Stephen R and Cole-Lewis, Heather and others},
  journal = {Nature Medicine},
  volume = {31},
  number = {3},
  pages = {943--950},
  year = {2025}
}

@inproceedings{shi2023distracted,
  title = {Large language models can be easily distracted by irrelevant context},
  author = {Shi, Freda and Chen, Xinyun and Misra, Kanishka and Scales, Nathan and Dohan, David and Chi, Ed and Sch{\"a}rli, Nathanael and Zhou, Denny},
  booktitle = {International Conference on Machine Learning (ICML)},
  pages = {31210--31227},
  year = {2023}
}

@article{coussement2024explainable,
  title = {Explainable AI for enhanced decision-making},
  author = {Coussement, Kristof and Abedin, Mohammad Zoynul and Kraus, Mathias and Maldonado, Sebasti{\'a}n and Topuz, Kazim},
  journal = {Decision Support Systems},
  volume = {184},
  pages = {114276},
  year = {2024}
}

@inproceedings{karpukhin2020dpr,
  title = {Dense passage retrieval for open-domain question answering},
  author = {Karpukhin, Vladimir and Oguz, Barlas and Min, Sewon and Lewis, Patrick and Wu, Ledell and Edunov, Sergey and Chen, Danqi and Yih, Wen-tau},
  booktitle = {Proceedings of the 2020 Conference on Empirical Methods in Natural Language Processing (EMNLP)},
  pages = {6769--6781},
  year = {2020}
}

@article{ada2025,
  title = {Standards of Care in Diabetes},
  author = {American Diabetes Association},
  journal = {Diabetes Care},
  volume = {48},
  number = {Suppl. 1},
  pages = {S1--S387},
  year = {2025}
}

@article{huang2025survey,
  title = {A survey on hallucination in large language models: Principles, taxonomy, challenges, and open questions},
  author = {Huang, Lei and Yu, Weijiang and Ma, Weitao and others},
  journal = {ACM Transactions on Information Systems},
  volume = {43},
  number = {2},
  pages = {1--55},
  year = {2025}
}

@inproceedings{zhao2025medrag,
  title = {MedRAG: Enhancing retrieval-augmented generation with knowledge graph-elicited reasoning for healthcare copilot},
  author = {Zhao, Xiangru and others},
  booktitle = {Proceedings of the ACM Web Conference 2025},
  pages = {4442--4457},
  year = {2025}
}

@article{jin2023medcpt,
  title = {MedCPT: Contrastive pre-trained transformers with large-scale PubMed search logs for zero-shot biomedical information retrieval},
  author = {Jin, Qiao and Kim, Woojeong and Chen, Qingyu and others},
  journal = {Bioinformatics},
  volume = {39},
  pages = {btad651},
  year = {2023}
}

@misc{openai2024gpt4,
  title = {GPT-4 Technical Report},
  author = {OpenAI},
  year = {2024},
  howpublished = {\url{https://arxiv.org/abs/2303.08774}}
}

@misc{llamaindex2023,
  title = {LlamaIndex: Data Framework for LLM Applications},
  author = {Liu, Jerry},
  year = {2023},
  howpublished = {\url{https://github.com/run-llama/llama_index}}
}

@article{singhal2023med,
  title = {Large language models encode clinical knowledge},
  author = {Singhal, Karan and Azizi, Shekoofeh and Tu, Tao and Mahdavi, Sara S and Lau, Jonathan and Barnett, Jacob C and Bifulco, Cesar and Callahan, Andrew and Chang, Nancy and Gentzel, Carolyn and others},
  journal = {Nature},
  volume = {620},
  pages = {172--180},
  year = {2023}
}

@article{guyatt2011grade,
  title = {GRADE: an emerging consensus on rating quality of evidence and strength of recommendations},
  author = {Guyatt, Gordon and Oxman, Andrew D and Vist, Gunn E and Kunz, Regina and Falck-Ytter, Yngve and Alonso-Coello, Pablo and Schünemann, Holger J},
  journal = {BMJ},
  volume = {336},
  number = {7650},
  pages = {924--926},
  year = {2011}
}

@article{kwan2020meta,
  title = {Computerised clinical decision support systems and absolute improvements in care: meta-analysis of controlled clinical trials},
  author = {Kwan, Jessica L and Lo, Linda and Ferguson, Jane and Ghali, William A and Rabi, Diane},
  journal = {BMJ},
  volume = {359},
  pages = {j4437},
  year = {2020}
}

@article{chen2023harnessing,
  title = {Harnessing the power of clinical decision support systems: challenges and opportunities},
  author = {Chen, Z and others},
  journal = {Open Heart},
  volume = {10},
  number = {1},
  pages = {e001878},
  year = {2023}
}

@article{aljohani2025trustworthiness,
  title = {A comprehensive survey on the trustworthiness of large language models in healthcare},
  author = {Aljohani, Manar and Hou, Jun and Kommu, Sindhura and Wang, Xuan},
  journal = {npj Digital Medicine},
  volume = {8},
  pages = {1--18},
  year = {2025}
}

@article{future2024,
  title = {FUTURE-AI: Guiding principles and consensus recommendations for responsible development and deployment of artificial intelligence in healthcare},
  author = {Holzinger, Andreas and Langs, Georg and Denk, Helmut and others},
  journal = {BMJ Health Care Informatics},
  volume = {31},
  number = {1},
  pages = {e100623},
  year = {2024}
}

\end{document}